\documentclass[sigconf]{acmart}

\AtBeginDocument{%
  \providecommand\BibTeX{{%
    \normalfont B\kern-0.5em{\scshape i\kern-0.25em b}\kern-0.8em\TeX}}}

\usepackage[ruled,vlined]{algorithm2e}

\setcopyright{acmcopyright}
\copyrightyear{2022}
\acmYear{2022}
\acmDOI{XXXXXXX.XXXXXXX}
\acmConference[Conference acronym 'XX]{Make sure to enter the correct
  conference title from your rights confirmation emai}{June 03--05,
  2022}{Woodstock, NY}
\acmPrice{15.00}
\acmISBN{978-1-4503-XXXX-X/22/06}

\begin{document}

\title{Hierarchical Capsule Prediction Network for Marketing Campaigns Effect}

\author{Zhixuan Chu}
\authornote{Correspondence to chuzhixuan.czx@alibaba-inc.com.}
\author{Hui Ding}
\author{Guang Zeng}
\affiliation{%
  \institution{Ant Group}
  \city{Hangzhou}
  \country{China}
  }

\author{Yuchen Huang}
\author{Tan Yan}
\author{Yulin Kang}
\affiliation{%
  \institution{Ant Group}
  \city{Hangzhou}
  \country{China}
  }

\author{Sheng Li}
\affiliation{%
  \institution{University of Virginia}
  \city{Charlottesville}
  \country{USA}}

\renewcommand{\shortauthors}{Chu, et al.}

\begin{abstract}
Marketing campaigns are a set of strategic activities that can promote a business’s goal. The effect prediction for marketing campaigns in a real industrial scenario is very complex and challenging due to the fact that prior knowledge is often learned from observation data, without any intervention for the marketing campaign. Furthermore, each subject is always under the interference of several marketing campaigns simultaneously. Therefore, we cannot easily parse and evaluate the effect of a single marketing campaign. To the best of our knowledge, there are currently no effective methodologies to solve such a problem, i.e., modeling an individual-level prediction task based on a hierarchical structure with multiple intertwined events. In this paper, we provide an in-depth analysis of the underlying parse tree-like structure involved in the effect prediction task and we further establish a Hierarchical Capsule Prediction Network (HapNet) for predicting the effects of marketing campaigns. Extensive results based on both the synthetic data and real data demonstrate the superiority of our model over the state-of-the-art methods and show remarkable practicability in real industrial applications. 
\end{abstract}



\maketitle
\section{Introduction}
\label{intro}
The effect of marketing campaigns usually has a significant impact on the company's business operation and planning. Therefore, an accurate effect prediction for the upcoming marketing campaign is essential, which helps guide the corresponding business operations in advance. In fact, the effect prediction for marketing campaigns in a real industrial scenario is very complex and challenging \cite{lings1998implementing,levy2010facebook}. Unlike the experimental study where the researcher can undertake different randomized controlled experiments to estimate the effect of marketing campaigns, in industry, the researcher often simply utilizes the observational data and arrives at a conclusion without any intervention. In addition, each subject is always under the interference of several marketing campaigns simultaneously. The involved marketing campaigns are often intertwined. For example, a certain marketing campaign might enlarge or shrink the effect of another marketing campaign on the subjects. The influence they have on each other is extraordinarily tangled and complicated. According to the actual business requirements, the effect prediction for the marketing campaign is always facing three primary challenges.

Firstly, since a large number of marketing campaigns take effect every day, each user has a high probability of being influenced by multiple campaigns, simultaneously. In addition, the campaigns also affect each other. For example, the interest-discounted loan and loan limit increase can complement and reinforce each other. In such a case, we cannot understand the interaction of heterogeneous marketing campaigns or determine the specific contribution of each campaign. It is very difficult to quantify the effect of each individual marketing campaign and predict the overall effect of the combination of campaigns.

Secondly, besides the entanglement of simultaneous marketing campaigns, we also face the unfixed combination of marketing campaigns. Different users are always in various combinations of marketing campaigns. For instance, there exist three main marketing campaigns at the same time, i.e., $a$, $b$, and $c$. The subjects may be influenced by different campaign combinations, such as $a$, $b$, $c$, $a + b$, $a + c$, $b + c$, or $a + b +c$. How to effectively and flexibly make predictions based on the different combinations of marketing campaigns for an individual subject is very challenging.

Thirdly, unlike the conventional prediction task based on the direct relationship between feature variables and outcome, event effect prediction is actually a hierarchical regression task involving several different levels. For example, the informative message passes in a hierarchical structure (individual feature level $\rightarrow$ single event level $\rightarrow$ event combination level $\rightarrow$ outcome level). It is of particular importance to consider such progressive part-whole relationships.

For these aforementioned practical challenges in the industry, there are currently no effective methodologies to solve such problems, i.e., modeling an individual-level prediction task based on the hierarchical structure with intertwined events. In this paper, we provide an in-depth analysis of the underlying parse tree-like structure \cite{sabour2017dynamic} involved in the effect prediction task and we further propose a \textbf{H}ierarchical C\textbf{a}psule \textbf{P}rediction \textbf{Net}work (HapNet). Similar to the human visual system that relies on a parse tree-like structure to recognize objects \cite{sabour2017dynamic}, human cognition and prediction procedures are also based on the composition and evolution of things from a lower level to a higher level \cite{aly2019hierarchical,liu2020multi,bi2022hierarchical}. Specifically, given the individual feature of subjects, the effect of a single event is inferred, several involved events are integrated organically, and finally, the total effect of the combination of events is learned. Capsule neural network has proved its effectiveness in modeling such hierarchical relationships on image data \cite{sabour2017dynamic,hinton2018matrix}. Therefore, we extend the capsule neural networks to reason these hierarchical relationships between the part and the whole in the prediction task. We create a new pipeline for predicting the effects of marketing campaigns, including the disentangled feature capsule, event capsule, event cluster capsule, and final outcome capsule. Extensive results on the synthetic and real datasets demonstrate the superiority of our model over the state-of-the-art methods and also show the remarkable practicality of our model in real industrial scenarios. 

\section{Background}
\label{Background}

\subsection{Clarification on Problem Setup}
Our major task is to predict the effect of marketing campaigns with the consideration of complex interrelationships among multiple marketing campaigns. Compared with the conventional prediction task based on the direct relationship between feature variables and outcome, our event effect prediction is actually a hierarchical prediction task including several levels from a lower level to a higher level \cite{richardson2015hierarchical,villegas2017learning}. Furthermore, we also face another challenge that is not appeared in conventional prediction tasks, i.e., the unfixed event combination with tangled interrelationships. The experiments provided in Section \ref{results} also demonstrate that the existing regression models tend to underperform, such as XGBoost \cite{chen2016xgboost}, Lasso regression \cite{roth2004generalized}, fuzzy k-nearest neighbor regression model \cite{mailagaha2021generalized}, vanilla deep multi-layer neural network \cite{larochelle2009exploring}, support vector regression (SVR) \cite{awad2015support}, and Transformer \cite{vaswani2017attention}. It is necessary to design a model dedicated to solving such a multi-level and multi-event regression problem.

\subsection{Clarification on Causal Inference}

When it comes to the conventional effect estimation task, we always face the selection bias, which is a core challenge involved in the causal inference and treatment effect estimation task~\cite{rubin1974estimating,li2017matching,yao2020survey,chu2020matching,chu2021graph,chu2022learning}. In fact, there is also a selection bias in our case. These marketing campaigns are typically not assigned randomly to the users due to the nature of observational data. This marketing campaign assignment bias results in that the treated population may differ significantly from the general population. Therefore, in the treatment effect estimation task, it's necessary to estimate the counterfactual outcomes while mitigating the selection bias. However, although there exists the marketing campaign assignment bias for users, we do not need to deal with this selection bias in our task. Our objective is just to predict the user's effect under the actual marketing campaign assignment with the selection bias, rather than estimating the counterfactual effect under other marketing campaigns that were not really assigned. Therefore, this marketing campaign effect prediction task essentially is a regression task, which is different from the treatment effect estimation task in causal inference.

\subsection{Clarification on Multi-task Learning}
\label{multi-task}
The different marketing campaign combinations have different properties; thus, each shall be modeled independently. The multi-task learning would be the most appropriate technique for addressing this kind of problems~\cite{caruana1997multitask,ruder2017overview,chu2022multi}. However, in practical applications, multi-task learning based on different combinations is impossible to achieve. More formally, a combination is a selection of events from the event set that has distinct event members, i.e., a $k$-combination of a set $S$ is a subset of $k$ distinct elements of $S$. If the set has $n$ elements, the number of $k$-combinations is denoted as  $\mathcal{C}(n,k)$. Therefore, for $n$ marketing campaigns, there are multiple combinations including no marketing campaign $\mathcal{C}(n,0)$, single marketing campaign $\mathcal{C}(n,1)$, two marketing campaigns $\mathcal{C}(n,2)$, up to $n$ marketing campaigns $\mathcal{C}(n,n)$. Totally, there maybe exist $2^n$ combinations according to the following formula:
\begin{equation*}
    \mathcal{C}(n,0)+\mathcal{C}(n,1)+\mathcal{C}(n,2)+...+\mathcal{C}(n,n)=2^n
\end{equation*}
Correspondingly, if we adopt the standard multi-task learning framework, we need to establish $2^n$ sub-networks for all different marketing campaign combinations, which is impractical in large-scale industrial applications.

\begin{figure}[t]
  \centering
  \includegraphics[width=\linewidth]{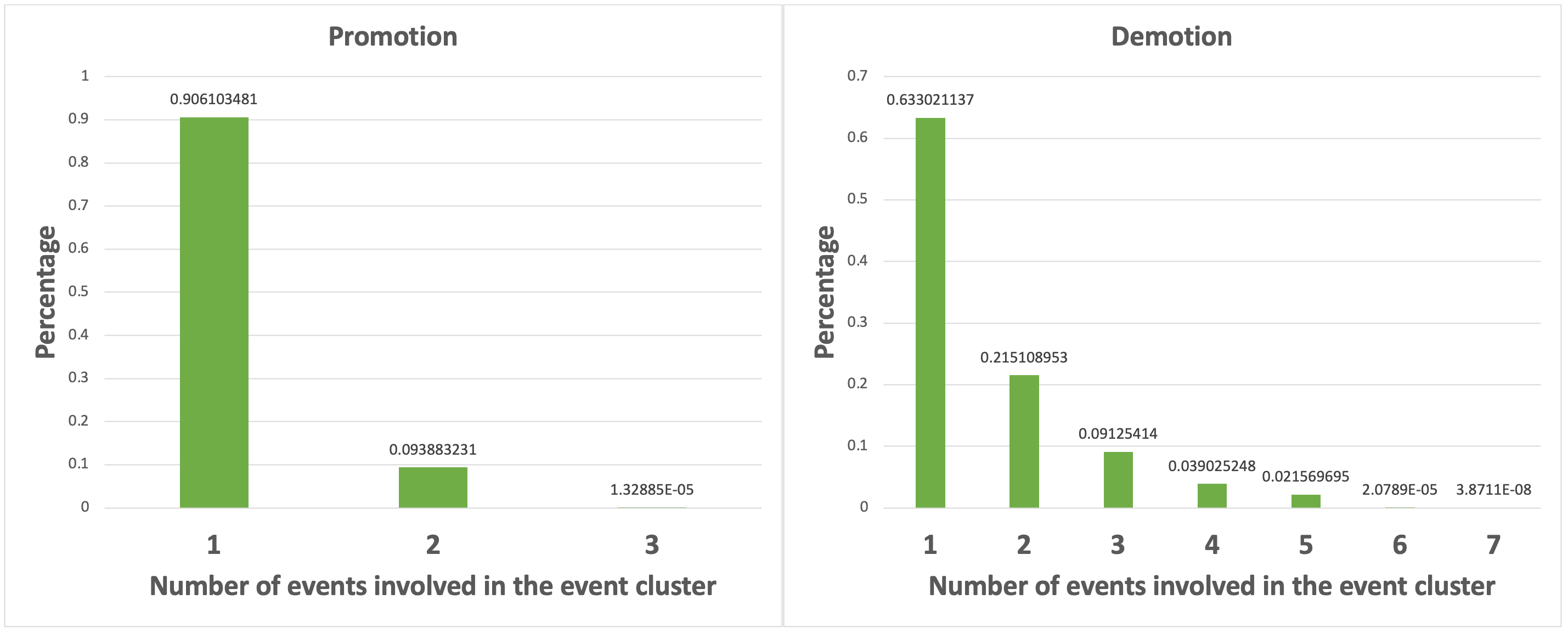}
  \caption{The distribution of user percentages influenced by the marketing campaign or marketing campaign combinations.}
  \label{dist}
\end{figure}

In addition, there is another challenge that multi-task learning cannot easily handle. As shown in Figure \ref{dist} obtained from the real data described in Section \ref{real_data}, the distributions of users influenced by the marketing campaigns or marketing campaign combinations are uneven. Most users are only influenced by one marketing campaign, and the number of users decreases as the number of events increases. It means that the marketing campaign combinations, especially with multiple marketing campaigns, do not have enough data to independently train a sub-network. In conclusion, it is infeasible to apply multi-task learning to this practical case of predicting the marketing campaign effect.

\section{The proposed framework}
\begin{figure*}[h]
  \centering
  \includegraphics[width=0.97\linewidth]{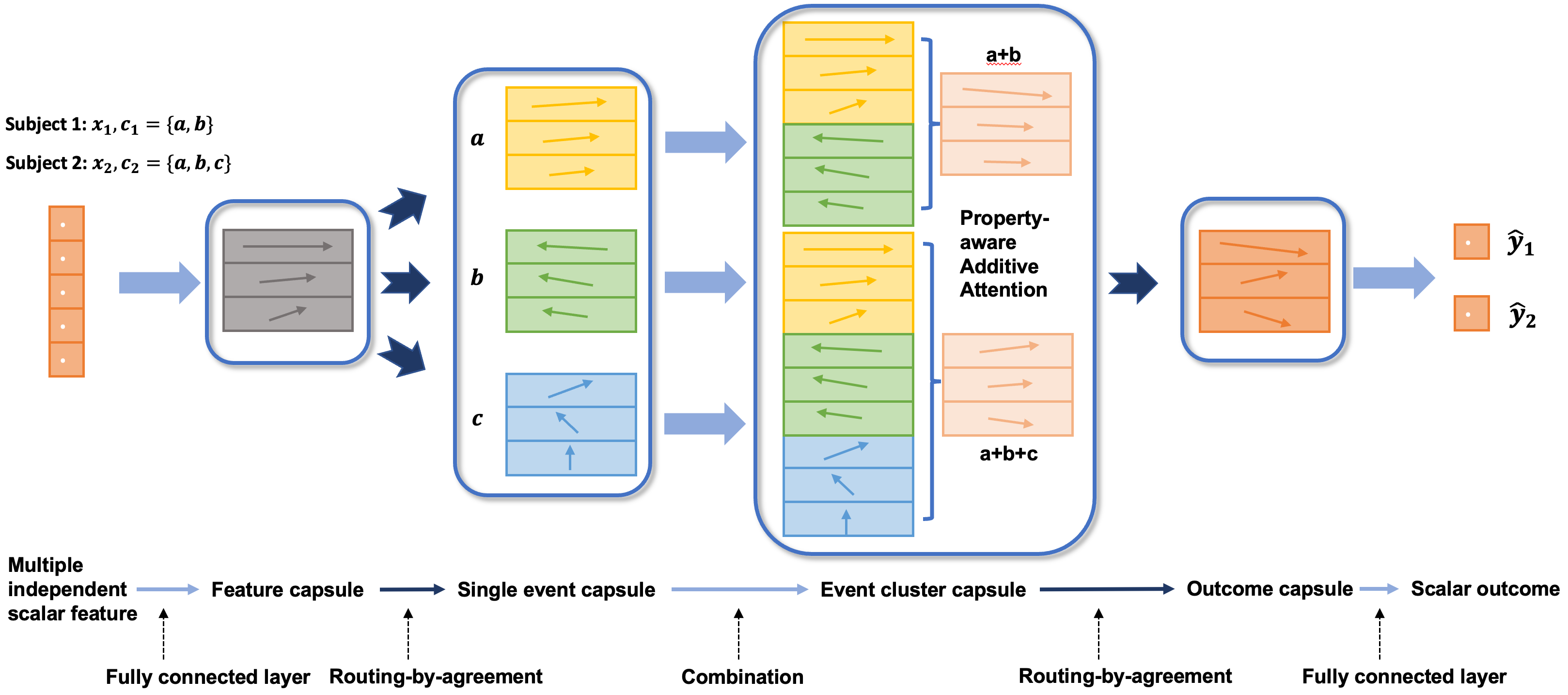}
  \caption{The framework of our HapNet model. It contains the disentangled feature capsule, single event capsule, event cluster capsule, and outcome capsule. The whole pipeline follows a hierarchical structure from part to whole. Here, we take two subjects with different marketing campaign combinations as examples.}
  \label{fig:framework}
\end{figure*}

As shown in Figure \ref{fig:framework}, our HapNet model follows a hierarchical structure from a lower level to a higher level, which contains the disentangled feature capsule, single event capsule, event cluster capsule, and outcome capsule. In the following, we will first present the problem formulation and then detail each component of our HapNet model.

\subsection{Problem Formulation}
The objective of our model is to predict the effect outcome of each subject under assigned marketing campaigns or the combination of marketing campaigns by figuring out the interrelationships among different marketing campaigns. More generally, we use ``events'' to refer to ``marketing campaigns'' throughout this paper.

Suppose that the observational data contain $N$ subjects and $n_e$ events. Let $\mathbf{x}_m \in \mathbb{R}^d$ and $y_m \in \mathbb{R}$ denotes observed feature variables and final effect outcome of subject $m; m=1,...,N$ under the influence of actual event or event cluster. The events $e; e=1,..,n_e$ can form $n_c$ event cluster, i.e., the observed combination of events. For example, for an observation ($\mathbf{x}_m,\, e = 1 \, \text{and} \,  3, \,y_m$), the event cluster $c$ of subject $m$ is $\{1,3\}$. We consistently focus on individual-level prediction, so for simplicity, we omit the subscript $m$ for the subject. The detailed notations are provided in Table \ref{tab:alg}.

\subsection{Disentangled Feature Capsule}

The general individual feature variables are considered as multiple independent scalar features, where each variable represents an attribute of a subject. Because highly complex interactions are involved in the connection from the individual feature level to the event level, the independent scalar features are proved to have limited capability in preserving the properties during the message passing \cite{verma2018graph,xinyi2018capsule,yang2020hierarchical}. In addition, the contribution to the effect of an event is not always from some single attribute, but it is often from an organic combination of several attributes. For example, we care about the socio-economic status of a subject, but we cannot directly measure that. However, the combination of some available attributes such as their zip code, income, and job type can be regarded as a proxy to represent the socio-economic status. Therefore, it is necessary to disentangle the explanatory factors underlying the feature attributes. Motivated by \cite{sabour2017dynamic}, we disentangle the latent factors of individual features and replace the scalar-based feature with vector-based feature capsules. In this way, the description of each subject is composed of multiple heterogeneous feature capsules and each feature capsule describes a specific instantiation parameter of the subjects. 

Formally, the feature of a subject is denoted by $\mathbf{x} \in \mathbb{R}^d$. Assuming that there exist $n_u$ latent explanatory factors, we project the input subject features into $n_u$ different spaces ($n_u$ feature capsules). The $i$-th feature capsule is defined as:
\begin{equation}
\hat{\mathbf{u}}_{i}=\sigma (\mathbf{W}_i^T \mathbf{x}) + \mathbf{b}_i,
\end{equation}
where $\mathbf{W}_i \in \mathbb{R}^{d \times h}$ and $\mathbf{b}_i \in \mathbb{R}^{h}$ are learnable parameters, $\sigma$ is a nonlinear activation function, and $h$ is the dimension of each feature capsule. Although more sophisticated implementations of feature disentanglement are possible \cite{ma2019disentangled}, we use linear projection in our study due to its efficiency and relatively remarkable performance. Recall that the existence probability of a capsule is measured by the capsule length \cite{sabour2017dynamic}, we use a non-linear {\bf "squashing"} function, which allows that the length of a short vector shrinks to almost zero and the length of a long vector tends to slightly below $1$ and we thus define that as follows:
\begin{equation}
{\mathbf{u}}_{i}= squash(\hat{\mathbf{u}}_{i}) = \frac{\|\hat{\mathbf{u}}_{i}\|^2}{1+\|\hat{\mathbf{u}}_{i}\|^2} \frac{\hat{\mathbf{u}}_{i}}{\|\hat{\mathbf{u}}_{i}\|},
\end{equation}
where $\mathbf{u}_i$ is the $i$-th feature capsule. Therefore, the feature capsules of each subject are represented by a pose matrix $\mathbf{U} = \{\mathbf{u}_{i}\}_{i=1}^{n_u} $. Therefore, the matrix $\mathbf{U}$ describes $n_u$ latent factors and is more informative in preserving the subject's properties than original features $\mathbf{x}$.

\subsection{Single Event Capsule}
To fully express the properties of the event from different perspectives, we still leverage the capsule, instead of conventional independent scalar features. In addition, to obtain a higher-level event capsule from the base-level individual feature capsule, it is essential to capture the part-whole relationship between adjacent capsule layers. Due to the heterogeneities of different events, the feature capsules may have different contributions to the different events. Furthermore, when the dimension of the feature capsule is high, directly concatenating the feature capsule and event indicators risks losing the influence of the event during training \cite{shalit2017estimating}. To combat this, we parameterize different networks for different events in the joint network. Therefore, we adopt a multi-task framework $\{f\}_1^{n_e}$, i.e., one dynamic routing \cite{sabour2017dynamic} for one event, where $n_e$ is the total number of potential events. Note that each subject is used to update only the networks corresponding to the actually involved events for that subject; for example, an observation ($\mathbf{x}, e = 1 \, \text{and} \,  3, \,y$) is only used to update $f_1$ and $f_3$. Similar to the feature capsules of subjects, the representation of events is composed of several event capsules, each of which describes one aspect of the properties. The fact that the output of a feature capsule is a vector makes it possible to use a powerful dynamic routing mechanism \cite{sabour2017dynamic} to ensure that the output of the feature capsule gets sent to an appropriate event capsule at the event level. 

The total input to a event capsule ${\hat{\bf s}}^{(e)}_j$ for event $e=1,...,n_e$ is a weighted sum over all ``prediction vectors'' ${\bf u}^{(e)}_{j|i}$ from the capsules in the individual feature level and ${\bf u}^{(e)}_{j|i}$ is produced by multiplying the output ${\bf u}_i$ by a weight matrix  ${\bf W}^{(e)}_{ij}$:
\begin{equation}
{\bf u}^{(e)}_{j|i} = {\bf W}^{(e)}_{ij}{\bf u}_i,
\end{equation}

\begin{equation}
{\hat{\bf s}}^{(e)}_j = \sum_i c^{(e)}_{ij} {\bf u}^{(e)}_{j|i},
\end{equation}
where  ${\bf W}^{(e)}_{ij} \in \mathbb{R}^{n_u \times n_s}$, $n_u$ is the number of feature capsules ,$n_s$ is the number of event capsules for each event, $i=1,...,n_u$, and $j=1,...,n_s$. The $c^{(e)}_{ij}$ are coupling coefficients that are determined by the iterative dynamic routing process, i.e., if the prediction vector has a large scalar product with the output of a possible parent, there is top-down feedback which increases the coupling coefficient for that parent and decreasing it for other parents \cite{sabour2017dynamic}. Each of these feature capsules is weighted by an routing weight $c^{(e)}_{ij}$ with which individual capsule is assigned to event capsule from part to whole, where $c^{(e)}_{ij} \geqslant 0$ and $\sum_{j=1}^{n_s} c^{(e)}_{ij}=1$. Formally, $c^{(e)}_{ij}$ is defined as $c^{(e)}_{ij} = {\text{exp}(b^{(e)}_{ij})}/{\sum_k \text{exp}(b^{(e)}_{ik})}$, where $b^{(e)}_{ij}$ is initialized as $b^{(e)}_{ij}=0$. In each iteration of the dynamic routing process, we have:
\begin{equation}
{\bf s}^{(e)}_j = squash({\hat{\bf s}}^{(e)}_j) = \frac{||{\hat{\bf s}}^{(e)}_j||^2}{1+||{\hat{\bf s}}^{(e)}_j||^2} \frac{{\hat{\bf s}}^{(e)}_j}{||{\hat{\bf s}}^{(e)}_j||}.
\label{squash}
\end{equation}
We update $b^{(e)}_{ij}$ with $b^{(e)}_{ij} = b^{(e)}_{ij} + a^{(e)}_{ij}$, where $a^{(e)}_{ij} = {\bf u}^{(e)}_{j|i} \cdot \mathbf{s}^{(e)}_j$ indicates the agreement between the current output ${\bf s}^{(e)}_j$ of the event capsule $j$ in the event level and the prediction ${\bf u}^{(e)}_{j|i}$ made by individual feature capsule $i$. Therefore, the event capsules for event $e$ are represented by a pose matrix $\mathbf{S}^{(e)} = \{\mathbf{s}^{(e)}_{j}\}_{j=1}^{n_s} $, where $n_s$ is the number of the event capsules for event $e$. Note that in our work, we set the number of the event capsules for all events to be the same, i.e., $n_s$. 

\subsection{Event Cluster Capsule}
After getting event capsules for a single event, we take the integrated effect of different events into account. In the real case, one subject always participates in multiple events and the interrelationship among different events are twisted and inexplicable. A certain event might enlarge or shrink the effect of another event on the subject. Therefore, above the event capsule level, there exists another higher-level event cluster capsule, which is responsible for the properties of a combination of events. According to the actual event cluster $c$ for the subject, we stack the event capsules ${\bf S}^{(e)}$ from the different event networks to attain the initial event cluster capsule ${\bf V}^{(c)}$, e.g., the event cluster $c$ for an observation ($\mathbf{x}, e = 1 \, \text{and} \,  3, \,y$) is $c=\{1,3\}$ and the initial event cluster capsule is ${\bf V}^{(c)} = \{{\bf S}^{(e)}\}_{e=1,3}$, where the event cluster capsule contains $n^{(c)}_v$ capsules, i.e., $n^{(c)}_v = n_{s_1}+n_{s_3}$.

\subsection{Property-aware Additive Attention}

\begin{figure}[t]
  \centering
  \includegraphics[width=\linewidth]{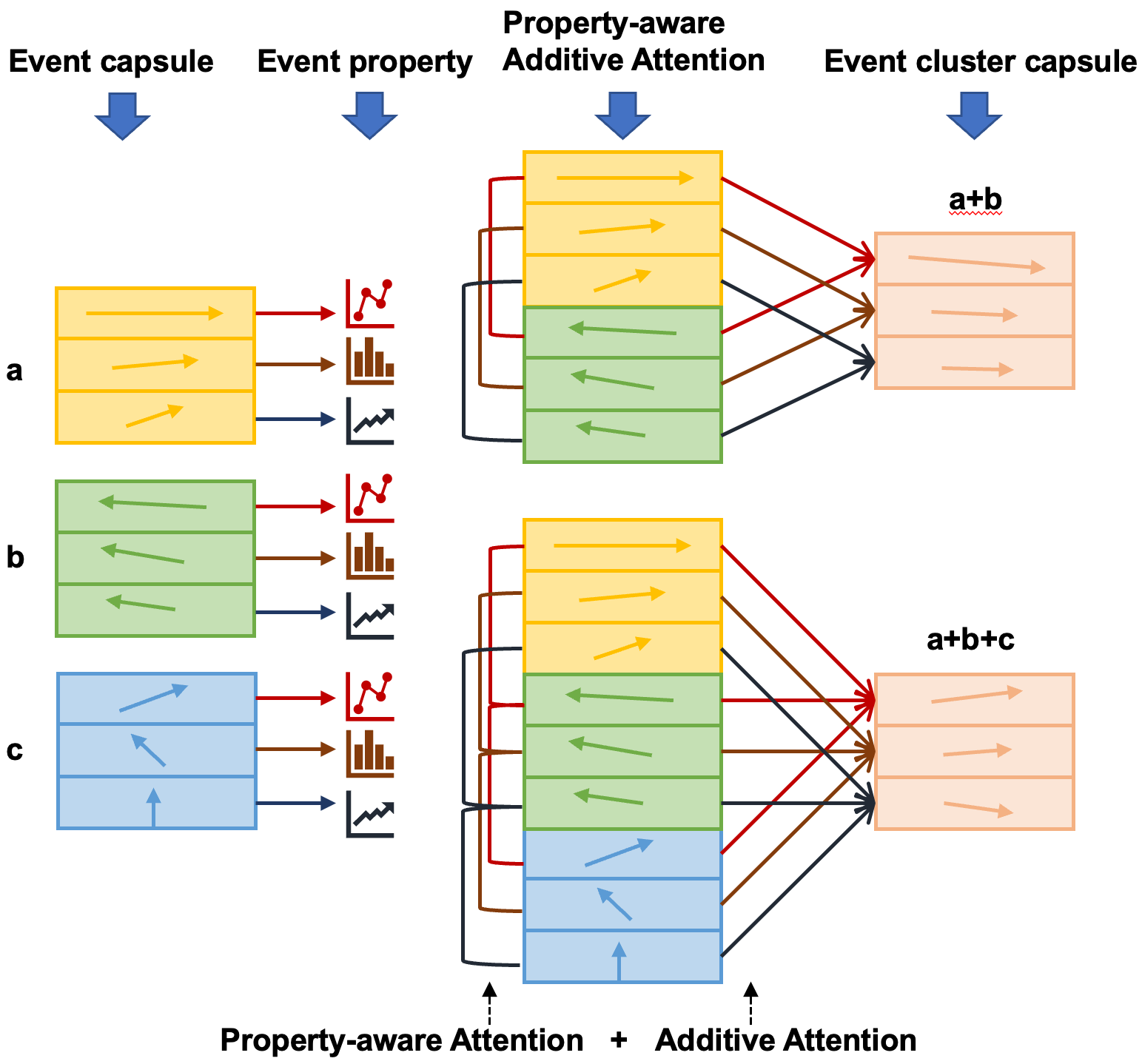}
  \caption{The procedure of property-aware additive attention. It contains property-aware attention and additive attention.}
  \label{fig:attention}
\end{figure}

Based on the initial event cluster stacked directly from event capsules, each initial event cluster contains different numbers of capsules (e.g., $\{\mathbf{x}, e = 1 \, \text{and} \,  3, \,y\}$, $c=\{1,3\}$,  $n^{(c)}_v=n_{s_1}+n_{s_3}$; $\{\mathbf{x}, e = 1, \,2, \, \text{and} \,  3, \,y\}$, $c=\{1,2,3\}$,  $n^{(c)}_v=n_{s_1}+n_{s_2}+n_{s_3}$). Thus, each event cluster has its unique characteristics. Naturally, compared to one fixed model, the multi-task framework for each event cluster is more reasonable. However, due to the large number of event combinations discussed in Section \ref{multi-task}, it is impossible to enumerate each combination on the basis of a multi-task framework. In addition, as shown in Figure \ref{dist}, the event clusters especially with multiple events do not have enough data to independently train a sub-network. It is imperative to design one framework that is appropriate for each event cluster with different numbers of events. Furthermore, to incorporate the consideration of interrelationships among different events in one event cluster, we design one property-aware additive attention module (Figure \ref{fig:attention}) before the dynamic routing between the event cluster capsule and the higher-level outcome capsule.

Due to the nature of the capsule network, each event capsule can describe one property of the corresponding event, for example, the event capsules ($n_s = 3$) can maybe separately represent the degree of event effect, the direction of event effect, and the range of event effect in the event level. Therefore, we aim to explore the interrelationship among the same property of the combined several events (the capsule at the same position in each event capsule). For each event cluster, we apply the attention mechanism to the capsules at the same position in each event capsule (e.g., for $\{\mathbf{x}, e = 1 \, \text{and} \,  3, \,y\}$, $c=\{1,3\}$, the stacked event cluster is $\Big\{ \{\mathbf{s}^{(e=1)}_{j}\}_{j=1}^{n_s}, \{\mathbf{s}^{(e=3)}_{j}\}_{j=1}^{n_s} \Big\}$. We apply attention to the $\{\mathbf{s}^{(e=1)}_{j=1}, \mathbf{s}^{(e=3)}_{j=1}\}$, $\{\mathbf{s}^{(e=1)}_{j=2}, \mathbf{s}^{(e=3)}_{j=2}\}$,.., $\{\mathbf{s}^{(e=1)}_{j=n_s}, \mathbf{s}^{(e=3)}_{j=n_s}\}$, respectively.) 

Here, for the simplicity of notations, we omit the index of event cluster $c$. The attention module function is parameterized with the following equation:
\begin{equation}
\bar{\bf s}_j^{(e=k)}=\sigma\left(\sum_{l} {\alpha_{kl} {\bf W} {\bf s}_j^{(e=l)}}\right),
\label{eq:attention}
\end{equation}
where $l$ is the event index in the event cluster $c$, which is used to loop through all event capsules at the position $j$ in the event cluster $c$. ${\bf W}$ is the learnable weight matrix and ${\bf W} {\bf s}_j^{(e=l)}$ is a linear transformation of ${\bf s}_j^{(e=l)}$. $\sigma(\cdot)$ is the activation function for nonlinearity. In Eq.~\eqref{eq:attention}, the representation of the $k$-th capsule and its neighbor capsules are aggregated together, scaled by the normalized attention scores $\alpha_{kl}$:
\begin{equation}
\alpha_{kl}=\frac{\exp(\text{LeakyReLU}( {a}^T({\bf W}{\bf s}_j^{(e=k)}||{\bf W}{\bf s}_j^{(e=l)})))}{\sum_{l}\exp(\text{LeakyReLU}( {a}^T({\bf s}_j^{(e=k)}||{\bf s}_j^{(e=l)})))},
\end{equation}
where softmax is used to normalize the attention scores on capsules at the same position of each event capsule in the same event cluster. The pair-wise attention score between two capsules is calculated by $\text{LeakyReLU}(a^T({\bf W}{\bf s}_j^{(e=k)}||{\bf W}{\bf s}_j^{(e=l)}))$. Here, it first concatenates the linear transformation of the representations for two capsules from two events, i.e., ${\bf W}{\bf s}_j^{(e=k)}||{\bf W}{\bf s}_j^{(e=l)}$,  where $||$ denotes concatenation, and then it takes a dot product of itself and a learnable weight vector $a$. Finally, the LeakyReLU function is applied. To stabilize the learning process, a multi-head attention mechanism based on the average operation is employed. 

To overcome the difficulty of different numbers of events to be combined, we then conduct a vector addition operation (adding the corresponding components of the vectors together) for the capsules at the same position of each event capsule in the same event cluster: 
\begin{equation}
 {\bf v}^{(c)}_j = \sum_l \bar{\bf s}_j^{(e=l)},
\end{equation}
where $l$ is the event index in the event cluster $c$. The event cluster capsule can be defined as ${\bf V}^{(c)} = \{{\bf v}^{(c)}_j\}_{j=1}^{n_{v}}$, where $n_{v}$ is the number of capsules and $j$ is the index of the capsule in the event cluster level (the same index as the single event level). Here, the number of event cluster capsules is equal to the number of event capsules, i.e., $n_{v} = n_s$, due to the mechanism of our designed property-aware additive attention. We note that for the subjects only influenced by a single event, we just conduct the linear transformation in the Eq. (\ref{eq:attention}) for the event capsule ${\bf s}_j^{(e)}$ to get the event cluster capsules with no need for property-aware additive attention. 

Similar to the ``routing-by-agreement'' between disentangled feature capsules and event capsules, we adopt the same strategy to learn the outcome capsules ${\bf Z}^{(c)} = \{{\bf z}^{(c)}_r\}_{r=1}^{n_z}$ from the attentional event cluster capsules ${{\bf V}}^{(c)} = \{{{\bf v}}^{(c)}_j\}_{j=1}^{n_{v}}$, where $n_{z}$ is the number of outcome capsules for the cluster $c$ and $r$ is the index of the outcome capsule. 

\subsection{Outcome Estimation}
Based on the above procedures, we can obtain learned outcome capsules ${\bf Z}_m$ for subject $m$ on the corresponding observed event or event cluster, we flatten the matrix ${\bf Z}_m$ into one outcome representation vector ${\bf \bar z}_m = Flatten({\bf Z}_m) $. Unlike the conventional classification task for the capsule network, we leverage a two-layer fully connected neural network $\psi({\bf \bar z}_m)$ to learn the one-dimensional scalar outcome $\hat{y}_m$ for each subject from flattened outcome representation vector. We aim to minimize the mean squared error in predicting the final outcomes of each subject on the corresponding event cluster:
\begin{equation}
\mathcal{L}_\Psi = \frac{1}{N}\sum_{m=1}^{N}(\hat{y}_m-y_m)^2,
\label{eq: factual}
\end{equation}
where $\hat{y}_m$ denotes the inferred observed outcome for subject $m$ corresponding to the factually observed event cluster. 

\begin{table*}[th!]
  \caption{The notation and algorithm of our HapNet. According to the procedure from Step $(1)$ to Step $(15)$, it achieves the pipeline, i.e., ``Feature $\rightarrow$ Feature capsule $\rightarrow$ Single Event capsule $\rightarrow$ Event cluster capsule $\rightarrow$ Outcome capsule $\rightarrow$ Outcome''. Here, the bold capital letter represents the matrix; the bold lowercase letter represents the vector; the regular lowercase letter indicates the scalar.}
  \label{tab:alg}
  \begin{tabular}{lccl}
    \toprule
    Name & Notation & Number & Comments \& Steps\\
    \midrule
    \textbf{Subject} & $m$ & $N$ & $m=1,...,N$\\
    \midrule
    \textbf{Feature of Subject} & $\mathbf{x}$ & $d$ & $\mathbf{x} \in \mathbb{R}^d$\\
    \midrule
    \textbf{Disentangled Feature Capsule} & $\mathbf{u}_{i}$ & $n_u$ & $i = 1,...,n_u$\\
    
    &  & & \textbf{(1)} $\hat{\mathbf{u}}_{i}=\sigma (\mathbf{W}_i^T \mathbf{x}) + \mathbf{b}_i$\\
    
    & & & \textbf{(2)}  ${\mathbf{u}}_{i}= squash(\hat{\mathbf{u}}_{i}) = \frac{\|\hat{\mathbf{u}}_{i}\|^2}{1+\|\hat{\mathbf{u}}_{i}\|^2} \frac{\hat{\mathbf{u}}_{i}}{\|\hat{\mathbf{u}}_{i}\|}$\\
    
    \midrule
    \textbf{Event} & $e$ & $n_e$ & $e=1,...,n_e$ \\
    \midrule
    \textbf{Single Event Capsule} & $\mathbf{s}^{(e)}_{j}$ & $n_s$ & $j=1,...,n_s$\\
    
    &  &  & \textbf{(3)}  ${\bf u}^{(e)}_{j|i} = {\bf W}^{(e)}_{ij}{\bf u}_i$\\
    
      &  &  & \textbf{(4)}  ${\hat{\bf s}}^{(e)}_j = \sum_i c^{(e)}_{ij} {\bf u}^{(e)}_{j|i}$\\
   &  &  & \textbf{(5)}  ${\bf s}^{(e)}_j = squash({\hat{\bf s}}^{(e)}_j) = \frac{||{\hat{\bf s}}^{(e)}_j||^2}{1+||{\hat{\bf s}}^{(e)}_j||^2} \frac{{\hat{\bf s}}^{(e)}_j}{||{\hat{\bf s}}^{(e)}_j||}$\\
  &  &  &  \textbf{(6)}  $\mathbf{S}^{(e)} = \{\mathbf{s}^{(e)}_{j}\}_{j=1}^{n_s} $\\

    \midrule
    \textbf{Event Cluster} & $c$ & $n_c$ &  e.g. for observation ($\mathbf{x}, e = 1 \, \text{and} \,  3, \,y$), $c=\{1,3\}$ \\
    
    \midrule
    \textbf{Event Cluster Capsule} & ${\bf v}^{(c)}_j$ & $n_{v}$ &  $j=1,...,n_{v}$\\
    
     & &  &  $\textbf{(7)}$ $\bar{\bf s}_j^{(e=k)}=\sigma\left(\sum_{l} {\alpha_{kl} {\bf W} {\bf s}_j^{(e=l)}}\right)$\\

      & & & \textbf{(8)}  ${\bf v}^{(c)}_j = \sum_l \bar{\bf s}_j^{(e=l)}$\\

     & & & \textbf{(9)}  ${{\bf V}}^{(c)} = \{{\bf v}^{(c)}_j\}_{j=1}^{n_{v}}$\\

    \midrule
    \textbf{Outcome Capsule} & ${\bf z}^{(c)}_r$ & $n_z$  & $r=1,...,n_z$\\
    
      & &  & \textbf{(10)}  ${{\bf v}}^{(c)}_{r|j} = {\bf W}^{(c)}_{jr} {{\bf v}}^{(c)}_j$\\
    
    & & & \textbf{(11)}  ${\hat{\bf z}}^{(c)}_r = \sum_j c^{(e)}_{jr} {{\bf v}}^{(c)}_{r|j}$\\
     
      &  &   &  \textbf{(12)}  ${\bf z}^{(c)}_r = squash({\hat{\bf z}}^{(c)}_r) = \frac{||{\hat{\bf z}}^{(c)}_r||^2}{1+||{\hat{\bf z}}^{(c)}_r||^2} \frac{{\hat{\bf z}}^{(c)}_r}{||{\hat{\bf z}}^{(c)}_r||}$\\
      
     &  &   & \textbf{(13)}  ${\bf Z}^{(c)} = \{{\bf z}^{(c)}_r\}_{r=1}^{n_z}$\\

    \midrule
    \textbf{Inferred Outcome} & $\hat{y}_m$ & 1 &  $\hat{y}_m \in \mathbb{R}^1$\\
    
     &  &  &  \textbf{(14)}  ${\bf \bar z}_m = Flatten({\bf Z}_m) $ \\
     &  &  & \textbf{(15)}  $\hat{y}_m = \psi({\bf \bar z}_m)$\\

  \bottomrule
\end{tabular}
\end{table*}

\subsection{Feature Reconstruction}
To encourage a series of capsules (Feature $\rightarrow$ Feature capsule $\rightarrow$ Event capsule $\rightarrow$ Event cluster capsule $\rightarrow$ Outcome capsule $\rightarrow$ Outcome) to exactly encode the instantiation parameters of the individual, event, event cluster, and outcome, we introduce a reconstruction loss to make the flatten outcome representation vector, ${\bf \bar z}_m$, capable of restoring the original individual feature. The ${\bf \bar z}_m$ is fed into a decoder consisting of several fully connected layers. We minimize the following feature reconstruction loss:
\begin{equation}
\mathcal{L}_\Phi = -\frac{1}{N}\sum_{m=1}^{N}\text{sim}({\bf \bar z}_m ,\mathbf{x}_m),
\label{eq: recon}
\end{equation}
where $-1 \leq \text{sim}(\cdot) \leq 1$ is the similarity between the inferred outcome representation vector ${\bf \bar z}_m$ and ground truth individual feature vector $\mathbf{x}_m$, respectively. $N$ is the total subject size. Intuitively, this loss encourages the model to produce high probabilities of reconstruction for the original feature vector. In the present experiments, we use cosine as the similarity measure. Therefore, putting all the above loss functions together, the objective function of our model is:
\begin{equation}
\mathcal{L} = \mathcal{L}_\Psi + \beta \mathcal{L}_\Phi,
\label{eq: total_loss}
\end{equation}
where $\beta \geq 0$ denotes the hyper-parameter controlling the trade-off between the outcome prediction loss $\mathcal{L}_\Psi $ and the reconstruction term $\mathcal{L}_\Phi$ in the final objective function. The notations involved in the model and the complete training process are provided in Table~\ref{tab:alg}.

\section{Experiments}

\subsection{Synthetic Dataset}
\begin{figure}
  \centering
  \includegraphics[width=0.8\linewidth]{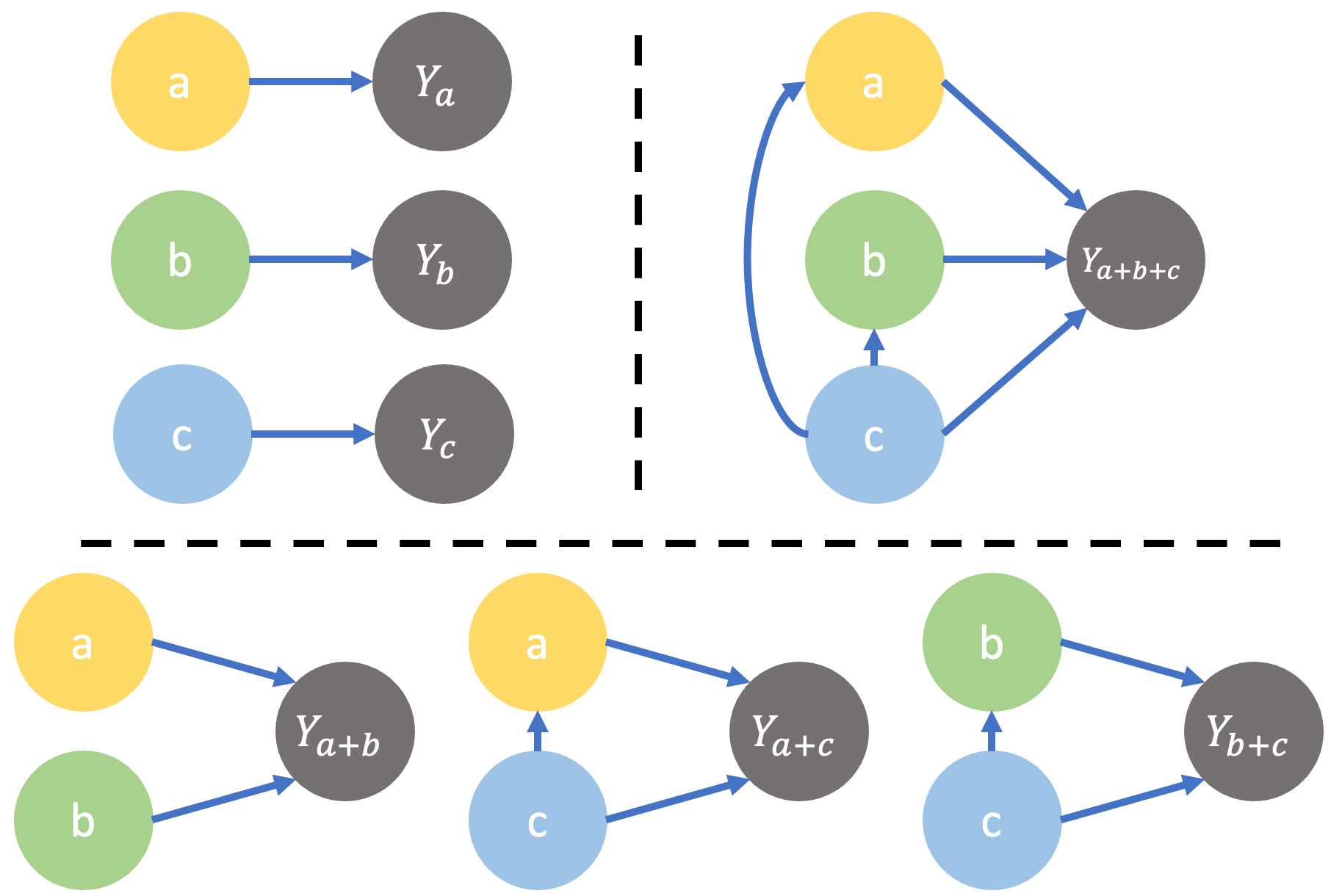}
  \caption{The interrelation among different events.}
  \label{fig:interrelations}
\end{figure}

Because in the observational data, we cannot intervene the event assignments, we cannot exactly observe the effect of each event or event cluster. We generate a synthetic dataset that can not only reflect the complexity of real data but also provide the exact estimation of each event and event cluster. For example, our synthetic data includes $3$ events, i.e., $a$, $b$, and $c$. Correspondingly, there are $7$ distinct event clusters, i.e., $a$, $b$, $c$, $a+b$, $a+c$, $b+c$, and $a+b+c$. In each event cluster, the same number of subjects are generated. One example of interrelations among these events is illustrated in Fig.~\ref{fig:interrelations}. When they are in the same event clusters, event $a$ and event $b$ are independent of each other and the event $c$ has a significant impact on event $a$ and event $b$, respectively. The detailed simulation procedures are provided:

\begin{align}
& y_a = \tau_a({\bf ax})+ \epsilon; \,\,\,\,\, y_b = \tau_b({\bf bx})+ \epsilon; \,\,\,\,\,  y_c = \tau_c({\bf cx})+ \epsilon; \\
& y_{a+b} = \tau_a({\bf ax})+ \tau_b({\bf bx}) + \epsilon; \\
& y_{a+c} = \tau_a({\bf (a+ \beta_{c\rightarrow a} )x }) + \tau_c({\bf cx }) + \epsilon; \\
& y_{b+c} = \tau_b({\bf (b + \beta_{c\rightarrow b} )x })+\tau_c({\bf  cx })+ \epsilon; \\
& y_{a+b+c} = \tau_a({\bf (a+\beta_{c\rightarrow a})x}) +\tau_b({\bf (b + \beta_{c\rightarrow b})x}) +\tau_c({\bf cx }) + \epsilon,
\label{Eqn: sim}
\end{align}
where ${\bf a}$, ${\bf b}$, and ${\bf c}$ are the effect coefficients for event $a$, $b$, and $c$, generated from $\text{Uniform}(-1,1)$. The $\beta_{c\rightarrow a}$ and $\beta_{c\rightarrow b}$ can represent the influences of event $c$ on event $a$ and event $b$, respectively. The $\tau_a$, $\tau_b$, and $\tau_c$ are different nonlinear trigonometric functions, which can describe the complexity of event effect estimation in real data. The vectors of all observed covariates ${\bf x}$ of subjects are sampled from a multivariate normal distribution. The noise $\epsilon$ is generated from the standard normal distribution. In order to explore the influence of event number, we generate $3$ synthetic datasets, i.e., \textbf{E3}, \textbf{E6}, and \textbf{E9} with $3$, $6$, and $9$ events, respectively. They all have similar interrelations as provided in Fig.~\ref{fig:interrelations}.

\subsection{Real Dataset}
\label{real_data}
To demonstrate the effectiveness of our method, we apply our model to the real datasets of Alipay, which is one of the world's largest mobile payment platforms and offers financial services to billion-scale users. Consistent with the challenges described in Section \ref{intro}, we have two different datasets (\textbf{Promotion} and \textbf{Demotion}) from two different scenarios, i.e., the one is to promote the activity of users with low credit risk and the other is to suppress activities of users with high credit risk. 

In the \textbf{Promotion} dataset, there are many special offers (such as free interest rates, an increase in loan limit, and so on) to promote a business's goal. In the \textbf{Demotion} dataset, for the company's sustainable and long-term development, they take some steps (such as an increase in the interest rate and appropriate reduction of loan limit) to control the loan amount of users who have a high risk of overdue loan repayment. The \textbf{Promotion} dataset contains millions of users, $6$ marketing campaigns, and $50$ feature variables of users. The \textbf{Demotion} dataset contains millions of users, $16$ activities, and $61$ feature variables of users. The $60\%$ and $20\%$ of the data are randomly taken for training and validation, respectively, and the remaining $20\%$ is used for testing.

\begin{table*}[ht]
  \caption{The mean and standard error of MAPE for both real datasets (Promotion and Demotion) and synthetic datasets (E3, E6, and E9). Best results are marked in bold (lower is better).}
  \label{tab:results}
  \begin{tabular}{cccccc}
    \toprule
    & \multicolumn{2}{c}{Real dataset} & \multicolumn{3}{c}{Simulation dataset}     \\
    \cmidrule(lr){2-3} \cmidrule(lr){4-6} 
    Method & \textbf{Promotion} & \textbf{Demotion} & \textbf{E3} & \textbf{E6} & \textbf{E9}\\

    \midrule
Lasso & $15.04\pm13.47$ & $20.15\pm19.10$ &  $7.80\pm5.82$ &  $10.55\pm9.27$ & $16.58\pm16.74$ \\
SVR & $11.81\pm10.31$ & $19.44\pm19.92$ &  $6.73\pm4.54$ &  $10.84\pm8.19$ & $15.12\pm14.99$ \\
DNN & $12.35\pm11.56$ & $18.27\pm17.79$ &  $5.29\pm4.90$ &  $9.15\pm7.95$ & $14.95\pm13.69$ \\
XGBoost & $10.76\pm9.10$ & $15.90\pm12.03$ &  $6.75\pm4.67$ &  $9.77\pm7.30$ & $12.01\pm10.07$ \\
FKNNR & $9.23\pm7.62$ &  $13.01\pm12.63$ &  $5.49\pm3.01$ & $8.82\pm6.71$ & $10.60\pm9.31$ \\
Transformer & $8.30\pm7.76$ & $13.44\pm11.45$ &  $5.64\pm3.78$ &  $8.36\pm6.68$ & $11.87\pm10.40$ \\
    \midrule
HapNet (w/o PAAA) & $7.25\pm3.98$ & $12.61\pm8.17$ &  $3.81\pm1.56$ &  $7.73\pm4.68$ & $10.35\pm8.48$ \\
HapNet (w/o RCON) & $6.30\pm4.73$ & $10.94\pm7.96$ &  $3.59\pm1.50$ &  $5.22\pm3.11$ & $7.47\pm6.35$ \\
HapNet (ours)& $\textbf{5.62}\pm \textbf{4.69}$ & $\textbf{10.20} \pm \textbf{7.73}$ &  $\textbf{3.26} \pm \textbf{1.47}$ &  $\textbf{4.60} \pm \textbf{2.55}$ & $\textbf{6.73} \pm \textbf{5.64}$ \\

  \bottomrule
\end{tabular}
\end{table*}

\subsection{Baseline Methods and Settings}

We compare our model with baseline models trained on synthetic and real datasets. According to the discussion in Section \ref{Background}, we apply some classical regression models to this task. XGBoost \cite{chen2016xgboost} is a sparsity-aware algorithm for sparse data and a weighted quantile sketch for approximate tree learning. Lasso regression \cite{roth2004generalized} is an efficient linear regression with shrinkage (Lasso). FKNNR \cite{mailagaha2021generalized} is a fuzzy k-nearest neighbor regression model based on the usage of the Minkowski distance instead of the Euclidean distance. DNN \cite{larochelle2009exploring} is a deep multi-layer neural network that has many levels of non-linearities allowing them to accurately represent complex non-linear functions. Support vector regression (SVR) \cite{awad2015support} is a type of support vector machine that supports linear and non-linear regression. Transformer \cite{vaswani2017attention} is a deep learning model that uses self-attention, differentially weighting the importance of each element of the input data. 

For our HapNet model, we set the number of disentangled feature capsule $n_u$, event capsule $n_s$, event cluster capsule $n_{v}$, outcome capsule $n_z$ to be the same i.e., $n_u =n_s= n_{v}=n_z=5$. The dimension of all capsules is set to $8$. The number of iterations in routing is set to $3$. The number of multi-heads in the property-aware additive attention is $3$. We scale the loss of feature reconstruction with $0.1$ ($\beta=0.1$) so that the model focuses on the prediction task.

\subsection{Results}
\label{results}
For both real datasets (Promotion and Demotion) and simulation datasets (E3, E6, and E9), we employ the mean absolute percentage error (MAPE) to measure the prediction accuracy of baseline models and our proposed HapNet model. It usually expresses the accuracy as a ratio defined by the formula: $\text{MAPE} = \frac{100\%}{N} \sum_{m=1}^N |\frac{y_m-\hat{y}_m}{y_m}|$, where $\hat{y}_m$ is predicted value for subject $m$. As shown in Table \ref{tab:results}, we provide the mean and standard error of MAPE for $5$ datasets, i.e., Promotion, Demotion, E3, E6, and E9. Our proposed HapNet consistently outperforms the state-of-the-art baseline methods with respect to both the mean and standard error of MAPE in all cases. 

\begin{figure*}[th!]
  \centering
  \includegraphics[width=0.9\linewidth]{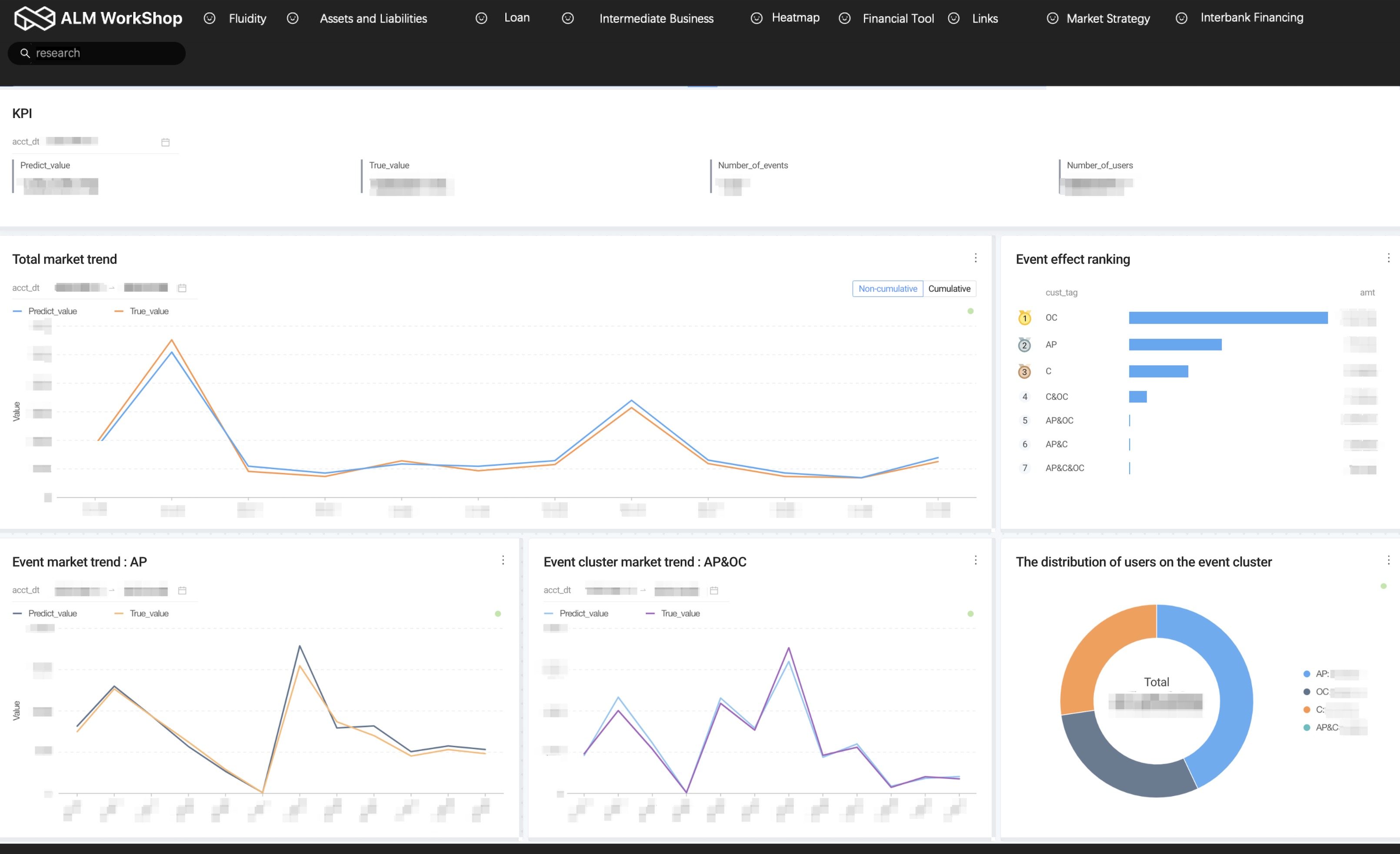}
  \caption{The deployment and result exhibition of HapNet. It presents different information in different modules, such as the prediction for total users (e.g., ``Total market trend''), the prediction for a group of people under the influence of a single marketing campaign (e.g., ``Event market trend: AP''), the prediction for a group of people under the influence of marketing campaign combination (e.g., ``Event cluster market trend: AP\&OC''). The exhibition system also includes some basic information, e.g., daily predicted value, true value, number of events, number of users, event effect ranking, and the distribution of users on the different events or event clusters.
}
  \label{deployment}
\end{figure*}

Furthermore, we aim to evaluate the effectiveness of the major components of the proposed HapNet model. We trained two ablation studies of HapNet (w/o PAAA) and HapNet (w/o RCON) on all datasets. The first one is HapNet (w/o PAAA) where the Property-aware Additive Attention (PAAA) is removed. We only conduct a vector addition operation (adding the corresponding components of the vectors) for the event capsules at the same position in the event cluster, to get the event cluster capsule. Therefore, the interrelationship among the events in the same event cluster is treated solely through vector addition. The second ablation study is HapNet (w/o RCON) where the feature reconstruction loss is removed and thus the model cannot make sure that the outcome representation vector is capable of restoring the original individual feature.

According to the results provided in Table \ref{tab:results}, the performance becomes poor after removing either the Property-aware Additive Attention or the feature reconstruction loss compared to the original HapNet. More specifically, after removing the Property-aware Additive Attention, the HapNet (w/o PAAA) suffers a dramatic decline in performance as the number of events involved increases, which means the PAAA is very helpful to handle the complex interrelation among events. Therefore, the Property-aware Additive Attention and the feature reconstruction are essential components of our model.

\section{Experimental Deployment}

In this section, we describe the deployment of our HapNet in the whole prediction system to handle the real data from Alipay, which is the top FinTech company in the world and offers financial services to billion-scale users. As the number of users grows, more and more marketing campaigns have been conducted to increase users' stickiness and promote business development. To provide decisions for financing plans in advance, the company needs to predict the effect of marketing campaigns. 

Our HapNet is used to predict the individual-level loan amount under the influence of intertwined marketing campaigns. Our model can utilize the data collected from the various external system to conduct the prediction and output the results to the result exhibition system. The result display page in the exhibition system is shown in Figure \ref{deployment}, which includes several different modules. Due to the fact that our model is based on individual-level prediction, we can easily obtain the event-level prediction, event cluster-level prediction, and prediction for total users. Therefore, different modules in the result display page can present results from different perspectives, such as the prediction for total users (e.g., ``Total market trend''), the prediction for a group of people under the influence of a single marketing campaign (e.g., ``Event market trend: AP''), the prediction for a group of people under the influence of marketing campaign combinations (e.g., ``Event cluster market trend: AP\&OC''). The exhibition system also includes some basic information, e.g., daily predicted value, true value, number of events, and number of users. In addition, it provides the event effect ranking and the distribution of users on the different events or event clusters. This result display page summarizes the effects of marketing campaigns from different perspectives.

\section{Conclusion}
The rapid development of Artificial Intelligence has enabled the deployment of various AI-based systems in industrial applications, but many of the AI algorithms being implemented nowadays fail to guarantee accuracy, robustness, and interpretability in the face of real and complex scenarios or settings. In this paper, for the common marketing campaigns effect prediction task in the industry, we propose a Hierarchical Capsule Prediction Network (HapNet) for modeling an individual-level prediction task based on a hierarchical structure with multiple intertwined events. HapNet utilizes the hierarchical capsule network and property-aware additive attention to effectively address the ``horizontal'' part-whole relationships from a lower level to a higher level and ``vertical'' unfixed event combinations. Extensive results based on both the synthetic and real data demonstrate the superiority of our model over the state-of-the-art methods, and show excellent practicability in real industrial applications.

\newpage
\bibliographystyle{ACM-Reference-Format}
\bibliography{sample-base}

\end{document}